\documentclass[10pt,twocolumn,letterpaper]{article}

\usepackage{wacv}
\usepackage{times}
\usepackage{epsfig}
\usepackage{graphicx}
\usepackage{amsmath}
\usepackage{amssymb}
\usepackage{algorithm}
\usepackage{algorithmic}
\usepackage{booktabs}

\wacvfinalcopy

\def\wacvPaperID{****} 

\ifwacvfinal\fi
\begin{document}

\title{Anomaly Detection Using GANs for Visual Inspection in Noisy Training Data}

\author{Masanari Kimura \\
Ridge-i Inc. \\
{\tt\small mkimura@ridge-i.com}
\and
Takashi Yanagihara \\
Ridge-i Inc. \\
{\tt\small tyanagihara@ridge-i.com}
}

\maketitle
\ifwacvfinal\thispagestyle{empty}\fi

\begin{abstract}
The detection and the quantification of anomalies in image data are critical tasks in industrial scenes such as detecting micro scratches on product. In recent years,  due to the difficulty of defining anomalies and the limit of correcting their labels, research on unsupervised anomaly detection using generative models has attracted attention.
Generally, in those studies, only normal images are used for training to model the distribution of normal images. The model measures the anomalies in the target images by reproducing the most similar images and scoring image patches indicating their fit to the learned distribution.
This approach is based on a strong presumption; the trained model should not be able to generate abnormal images. However, in reality, the model can generate abnormal images mainly due to noisy normal data which include small abnormal pixels, and such noise severely affects the accuracy of the model.
Therefore, we propose a novel anomaly detection method to distort the distribution of the model with existing abnormal images. The proposed method detects pixel-level micro anomalies with a high accuracy from $1024\times{1024}$ high resolution images which are actually used in an industrial scene.  In this paper, we share experimental results on open datasets, due to the confidentiality of the data.
\end{abstract}

\section{Introduction}
The detection and the quantification of anomalies in image data are critical tasks in many industries such as detecting micro scratches on product surfaces, or finding out diseases from medical images. There are many studies dealing with such tasks~\cite{quam1978road,kim2006image,an2015variational}.
These studies are addressing task-specific problems in detecting  anomalies from images.

For such tasks, applying supervised learning method is fairly hard in general due to the difficulty of defining anomalies and collecting enough number of abnormal data.
In recent years, research on unsupervised anomaly detection using generative models has attracted attention.

\begin{figure}
\includegraphics[viewport=150 70 425 842, scale=0.325, angle=270]{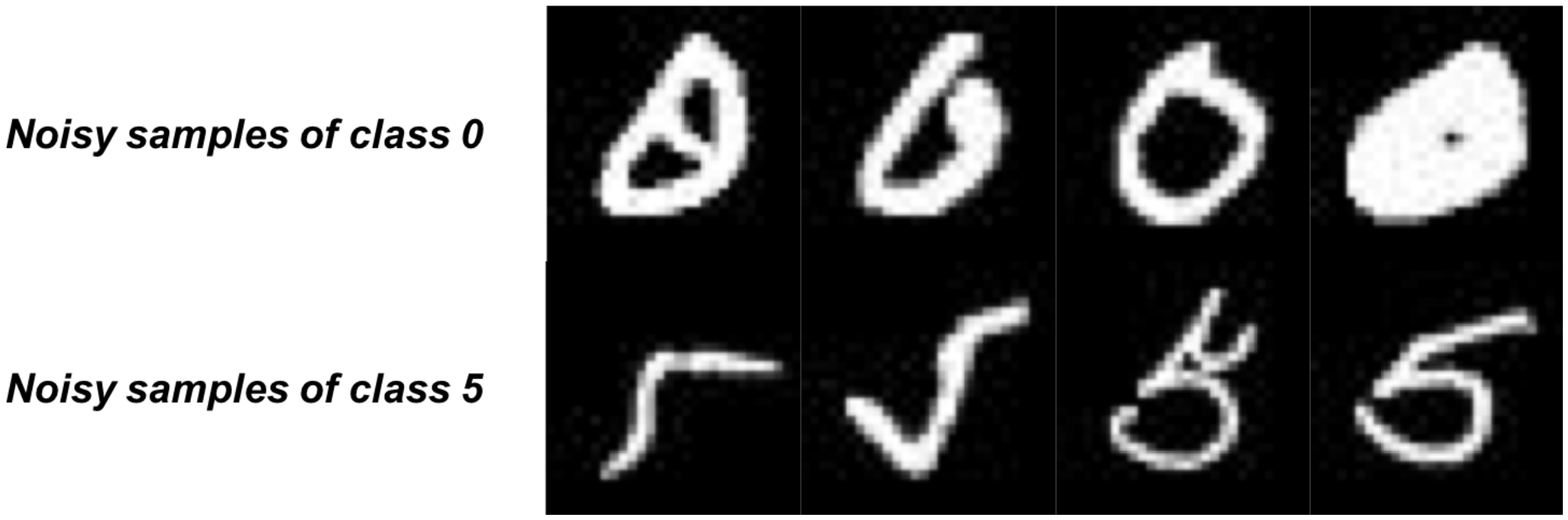}
\vspace{0.2cm} 
\caption{Noisy samples included in the MNIST handwritten digits dataset.}
\label{fig:mnist_noisy}
\end{figure}

 One of the most successful cases regarding generative model research is Generative Adversarial Networks (GANs)~\cite{goodfellow2014generative}. A GAN mimics the given target distribution by simultaneously training typically two networks, a generator $G$ and a discriminator $D$. 
The $G$ produces the model distribution  and $D$ distinguishes the model distribution from the target. 
This learning framework has been successful in various application fields such as image generation~\cite{pmlr-v48-reed16}, semantic segmentation~\cite{luc2016semantic},
image translation~\cite{zhu2017unpaired}\cite{yi2017dualgan}, and super resolution~\cite{ledig2017photo}, among others.
In this research, we apply GANs to anomaly detection.

There are several studies on anomaly detection using GANs~\cite{schlegl2017unsupervised, zenati2018efficient, wang2018generative, sabokrou2018adversarially}.
In those studies, only normal images are used to train GAN to model the distribution of the normal images. After the training is converged and a target image is queried, $G$ generates the most similar image to the target. When anomalies are included in the target image, there should be some distances between the target and the generated images, since the model only knows the distribution of the normal images. Whether the image is classified as normal or not is decided based on the threshold for the distance.
There is one common strong assumption among these approaches ; the model $G$ trained with normal images only should generate normal images. In other words, $G$ should not be able to generate abnormal images. 

\begin{figure*}[tbp]
\begin{center}
    \includegraphics[viewport=0 0 759 345, scale=0.6]{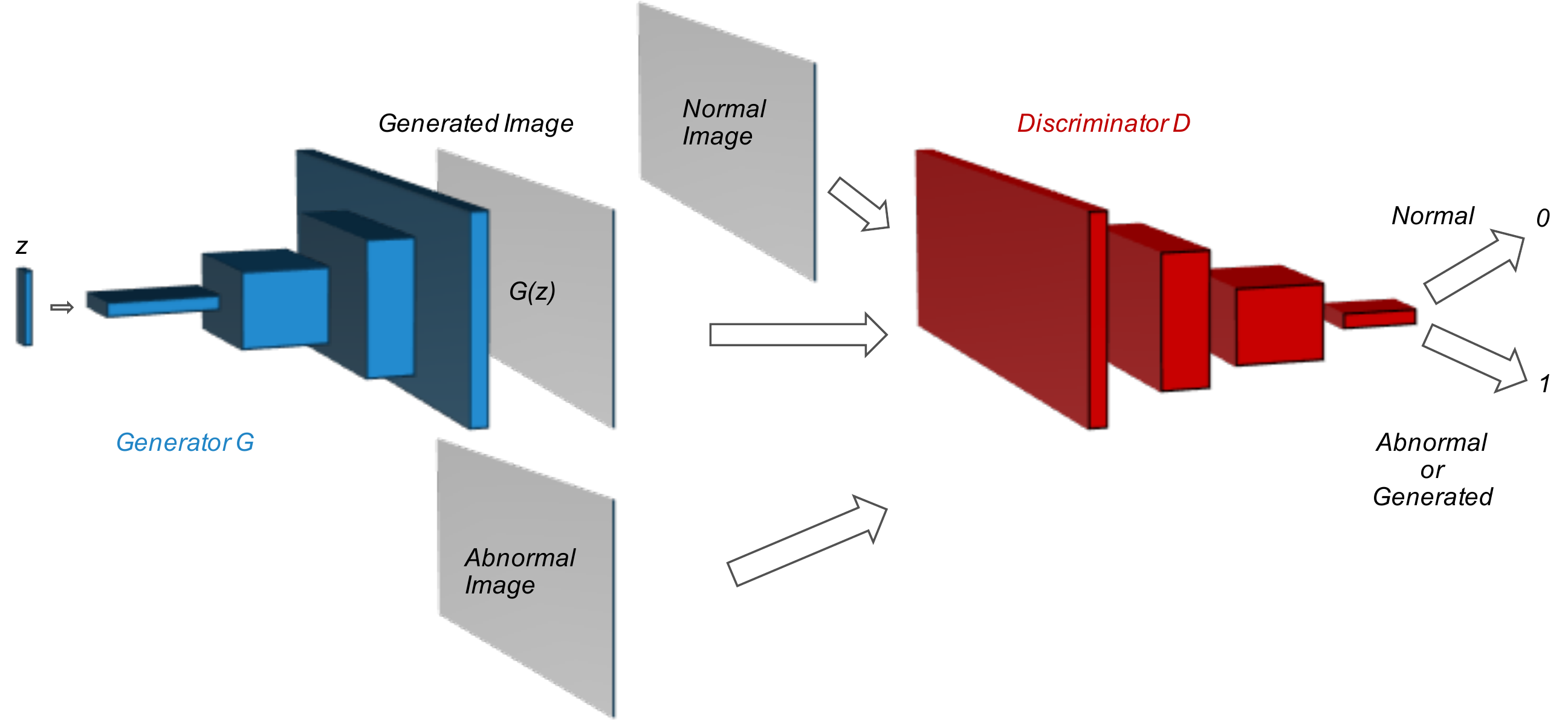}
\end{center}
\caption{Anomaly detection using Generative Adversarial Networks. A generator maps noise $z$ onto a real image. A discriminator classifies whether a target image is a normal image, or an abnormal or generated image. The discriminator conducts binary classification; 0 if the input image is normal and 1 if the input image is either abnormal or generated.}\label{fig:proposed_method}
\vspace{0.2cm}
\end{figure*}

However, in reality, the model can also generate abnormal images mainly due to a small number of abnormal pixels included in some normal images that are used for training. Trained with such noisy data, the generator recognizes them as part of the normal features. In real-world data, immaculate normal data are quite rare and it is virtually impossible to completely remove a few pixels of abnormal or distorted features included in normal images. (See Figure 1)


Therefore, we propose a anomaly detection method effectively utilizing given abnormal images to resolve the issue.
Our main contributions are the following:                                                                      
\begin{itemize}
\item{We solve the issues in the failure cases of the earlier studies on anomaly detection using GANs.}
\item{We propose a novel method for anomaly detection using GANs. Our method achieves accurate anomaly detection by utilizing both normal and abnormal images.}
\item{Our method successfully detects pixel-level micro anomalies in $1024\times{1024}$ high resolution images from an actual industrial scene.}
\end{itemize}
                                                                                                                                                                                                 
\section{Related Works}
In this section, we outline two studies referred in our research: Generative Adversarial Networks (GANs) and Anomaly Detection using GANs.
\subsection{Generative Adversarial Networks}
In recent years, GANs~\cite{goodfellow2014generative} have achieved a great success in image generation tasks.
A GAN consists of two networks, a generator $G$ and a discriminator $D$.
A generator learns the distribution $p_g$ by mapping noise $z$, which is sampled from the uniform distribution, to the image space $\chi$. A discriminator learns to distinguish between generated images and genuine images.
The discriminator and the generator are simultaneously optimized through the two-player minimax game as follows:
\begin{equation}
\label{eq:1}
\begin{split}
\min_{G}\max_{D} V(D, G) = \mathbb{E}_{x\sim{p_d(x)}}[\log{D(x)}] \\
 + \mathbb{E}_{x\sim{p_z(z)}}[\log{(1-D(G(z))}]
\end{split}
\end{equation}
Here, $p_{data}$ is the distribution of real data and $p_z$ is the distribution of noise $z$.
As adversarial training continues, the generator becomes able to generate samples that look similar to the real images, and the discriminator becomes able to identify whether an image is genuine or generated.

Despite of its successful performance in many fields, GANs' instability during the training has always been a critical issue, particularly with complicated images as in a photo-realistic high resolution case. 
To solve such problems and improve the learning stability of GANs, many studies have been released ~\cite{radford2015unsupervised, pmlr-v70-arjovsky17a,gulrajani2017improved,salimans2016improved,karras2018progressive} and Karras et al. proposed a method called progressive growing which gradually increases the resolution from low-resolution images throughout the learning phase~\cite{karras2018progressive}.
This method successfully generates much clearer high-resolution images compared to the existing methods can.
We applied the progressive growing framework to our research so that we can accurately detect small anomalies in high-resolution images.

\subsection{Anomaly Detection using GANs}
Recently GANs have been used in anomaly detection research and AnoGAN~\cite{schlegl2017unsupervised} brought a great progress to the field with a simple algorithm that only normal images are used for training a generator to model the distribution of normal images. 
With the trained $G$, if a given new query image $x$ is from the normal data distribution, noise $\hat{z}$ must exist in the latent space where $G(\hat{z})$ becomes identical to $x$. However if $x$ is abnormal, $\hat{z}$ will not exist even though $G$ tries to generate images most similar to $x$.  
The algorithm is heavily based on this hypothesis.

To find the noise $\hat{z}$, AnoGAN uses two loss functions : residual loss and discrimination loss.
\newline
\newline
{\bf Residual Loss} The residual loss measures the visual distance between the input image $x$ and the generated image $G(z)$.
\begin{equation}
\label{eq:2}
L_{R}(z) = \sum{|x - G(z)|}
\end{equation}
If $G$ perfectly learned the distribution of the normal data, $L_R$ should work as follows:
\begin{itemize}
\item{Input image is normal: $L_R \simeq 0$}
\item{Input image is abnormal: $L_R \gg 0$}
\end{itemize}
From the above, we can formulate visual differences.
\newline
\newline
{\bf Discrimination Loss based on feature matching} In addition to $L_{R}$, the discrimination loss is based on feature matching which uses an intermediate feature representation of the discriminator by 

\begin{equation}
\label{eq:3}
L_{D}(z) = \sum{|f(x) - f(G(z))|}
\end{equation}
where the output of the discriminator's intermediate layer $f(·)$ is used to extract the features of the input image and the generated image.
For $G$ to learn the mapping to the latent space,  overall loss is defined as the weighted sum of both components:
\begin{equation}
\label{eq:4}
L_{Ano}(z) = (1-\lambda)\cdot{L_{R}(z)} + \lambda \cdot{L_{D}(z)}
\end{equation}
With this differentiable loss function, $\hat{z}$ that makes the image generated by $G(\hat{z}$) most similar to the input image can be searched using back propagation.
The trained parameters of the generator $G$ and the discriminator $D$ are kept fixed during the search.
This is the fundamental mechanism of using GANs for anomaly detection.

\begin{figure*}[tbp]
\begin{center}
    \includegraphics[viewport=50 0 505 842, scale=0.5, angle=270]{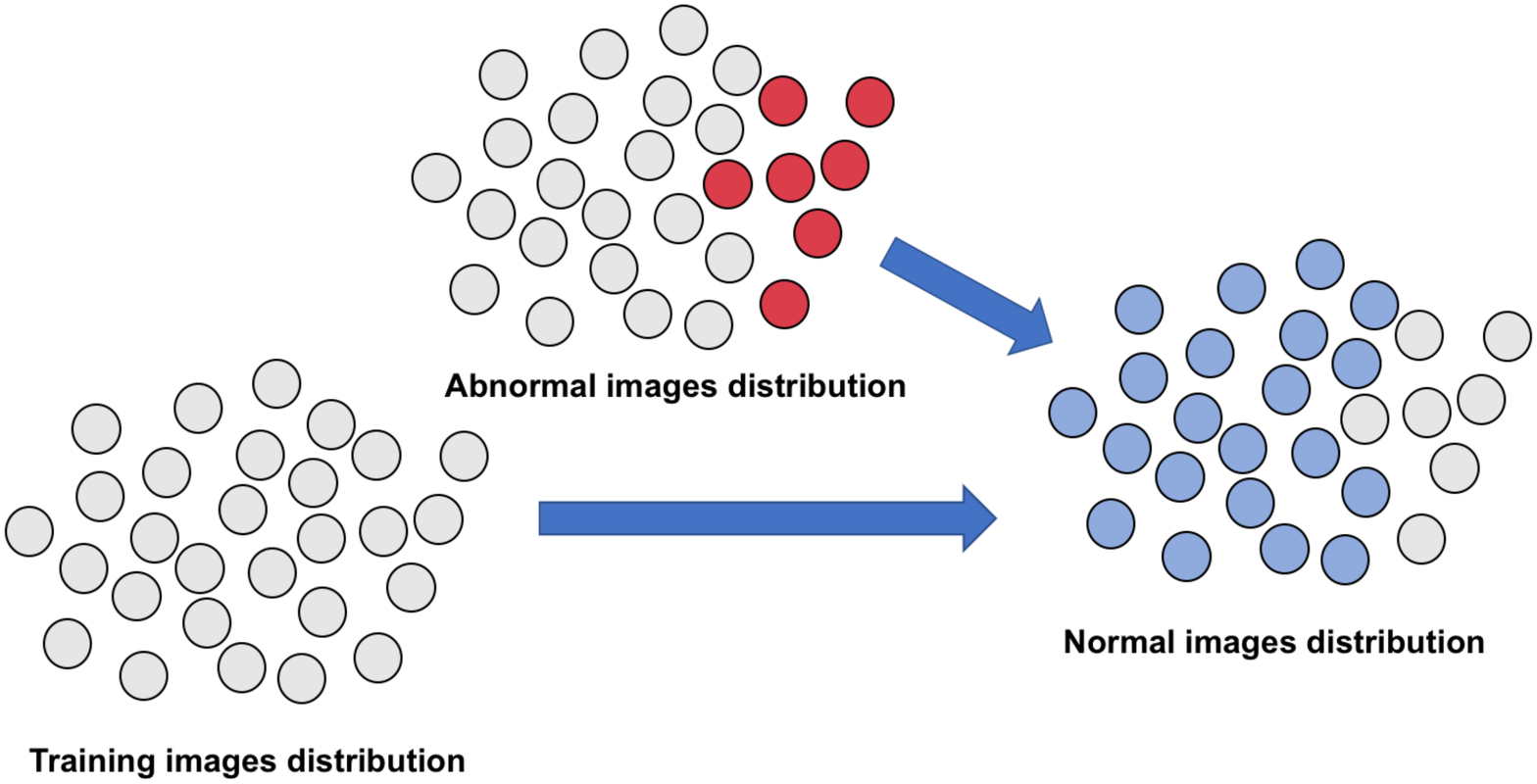}
\end{center}
\vspace{0.2cm} 
\caption{Visualization of distribution deformation by the proposed method. By using the known abnormal image set as much as possible, the proposed method excludes the distribution of abnormal images from the distribution that $G$ can take.}\label{fig:distortion}
\end{figure*}

\section{Proposed Method}
Using GANs for anomaly detection is based on a strong assumption that $G$ trained with normal images cannot generate abnormal images; $G$ should generate normal images only.
However, in reality, there are circumstances in which $G$ generates abnormal images due to the following factors:
\begin{itemize}
\item{The normal images used for training actually included some anomalies, and the generator learns it as normal features. Fig\ref{fig:mnist_noisy} shows the noisy samples included in the MNIST handwritten digits dataset.}
\item{The generator does not have enough representation power or enough training data to learn perfect mapping from $z$ to the image space.}
\end{itemize}
These are the natural behavior of the GAN architecture and we focus on dealing with unavoidable anomalies occurring in the dataset.
In real wild data, immaculate normal samples very rare and it is practically unfeasible to completely remove a few pixels of abnormal or distorted features found from the normal dataset.

To solve the problem, we propose a method of helping the generator learn distributions similar to the distribution of the normal data with  abnormal images that already exist. Fig \ref{fig:proposed_method} shows the overview of our method.

The proposed method reconstructs the learning framework of GANs.
The objective function $V(D, G)$ of GANs can be transformed from Equation~\ref{eq:1}  to the following equation:
\begin{equation}
\label{eq:5}
V(D, G) = \int_{x}{p_d(x)\log{D(x)} + p_{g} (x)\log{(1-D(x))}dx}
\end{equation}
Here, let $y = D(x)$, $a = p_d(x)$, $b = p_g(x)$,
\begin{eqnarray}
 \label{eq:6}
h(y) &=& a\log{y} + b\log{(1-y)} \\
 \label{eq:7}
\frac{d}{dy}h(y) &=& \frac{a}{y} + \frac{b}{y-1} \\
 \label{eq:8}
 &=& \frac{a(y-1) + b(y)}{y(y-1)} \\
 \label{eq:9}
 &=& \frac{y(a+b) - a}{y(y-1)}
\end{eqnarray}
Since $y = \frac{a}{a+b}$ when $\frac{d}{dy}h(y) = 0$ in Equation~\ref{eq:9}, the optimum discriminator is derived as below:
\begin{equation}
\label{eq:10}
D^{\ast}(x) = \frac{p_d(x)}{p_d(x) + p_g(x)}
\end{equation}
Now, we consider Jensen-Shannon divergence (JSD), which is defined as follows:
\begin{eqnarray}
\label{eq:11}
JSD(p_d||p_g) &\!=\!& \frac{1}{2}\!(\!KL(p_d||p_A) \!+\! KL(p_g||p_A)\!)\! \\
\label{eq:12}
KL(p_d||p_g) &=& \int_{x}{p_d(x)\log{\frac{p_d(x)}{p_g(x)}} dx} \\
\label{eq:13}
p_A &=& \frac{p_d+ p_g}{2}
\end{eqnarray}
By combining Equation~\ref{eq:5}, \ref{eq:10}, \ref{eq:11} and \ref{eq:12}, we obtain the optimum object function below:
\begin{equation}
\label{eq:14}
V(D^{\ast}, G) = 2JSD(p_d||p_g) - 2\log{2}
\end{equation}
From Equations~\ref{eq:14}, we can assume that the generator aims at minimizing the JSD between the real distribution and the generator distribution.

\begin{algorithm}[tbp]
\caption{Noisy-AnoGAN Inference}
\label{alg:1}
\begin{algorithmic}
\REQUIRE $x$, the input image, $n$, the number of iterations, $\gamma$, the weight of anomalies loss.
\STATE{Sample $z \sim p_z$}, a noise sample from the uniform distribution.
\FOR{$i=t\dots n$}
\STATE{$l\gets L'(z)$}
\STATE{$z\gets Update(z, l)$}
\ENDFOR
\STATE{$\hat{z}\gets z$}
\STATE{$anomaly score\gets L'(\hat{z})$}
\end{algorithmic}
\end{algorithm}

In contrast to the typical objective loss function -- Equation~\ref{eq:1} which only takes normal images into consideration, we define an additional loss function to consider abnormal images as well.
Our proposed method treats abnormal images as another type of generated images and adds penalty loss $l_{An}(D) $ with penalty weight $\gamma$ .
This can be regarded as distorting the data distribution $p_d$.
The objective loss function is defined as :
\begin{eqnarray}
\label{eq:15}
V'(D, G) &=& \gamma l_{Adv}(D, G) \!+\! (1\!-\!\gamma) l_{An}(D) ,
\end{eqnarray}
where
\begin{eqnarray}
l_{Adv}(D, G) &=& V(D, G) \\
l_{An}(D) &=& \mathbb{E}_{x\sim{p_{an}(x)}}[\log{(1-D(x)}] .
\end{eqnarray}
Here, $p_{an}$ is the distribution of abnormal images and $\gamma$ is a parameter of (0, 1].
The parameter $\gamma$ controls the percentage of abnormal images generated.
A smaller $\gamma$ excludes abnormal images from the training data, but at the same time it might even penalize normal features included in abnormal data.
Combining Equation~\ref{eq:14}, \ref{eq:15} and our proposed definition, we obtain the objective function of the generator as :
\begin{eqnarray}
V'(D^*, G) &=& 2JSD(p_d||p_N) - 2\log{2} \\
p_N &=& \gamma p_g + (1-\gamma) p_{an} .
\end{eqnarray}
Therefore, we can assume that the objective function of $G$ is to minimize the JSD between the real image distribution and the mixed distribution of generated images and abnormal images.
Since the JSD becomes the minimum when $p_d = p_N$, the optimum $p_g^*$ for $G$ can be derived as follows:
\begin{eqnarray}
p_N^* &=& p_d \\
\gamma p_g^* + (1-\gamma) p_{an} &=& p_d \\
\gamma p_g^* &=& p_d - (1-\gamma) p_{an} \\
p_g^* &=& \frac{1}{\gamma} p_d - \frac{(1-\gamma)}{\gamma} p_{an}
\end{eqnarray}

where $\frac{1}{\gamma}　\geq 1$ and $\frac{(1-\gamma)}{\gamma} = \frac{1}{\gamma} - 1 \geq 1$.
From the equations above, we may consider that the proposed method distorts the distribution of real images to remove abnormal images and to make the ideal distribution of normal images. Figure~\ref{fig:distortion} represents the change in the distribution caused by the proposed method.

During the inference, our method uses the following function to search the noise $\hat{z}$ that makes an image most similar to the input.
\begin{eqnarray}
L'(z) &=& \gamma L_{Ano}(z) + (1-\gamma) L'_{An}(z) \\
L'_{An}(z) &=& \sum{|1-D(G(z))|}
\end{eqnarray}

The function $Update(z, l)$ updates the noise $z$ based on the value of $l$. Algorithm~\ref{alg:1} shows the inference algorithm.

After finding the noise $\hat{z}$, we can use $anomalyscore$ $L'(\hat{z})$ to classify images in to normal and abnormal. 
Furthermore, we can identify the abnormal pixels in the image by calculating the difference between the generated image $G(\hat{z})$ and the input.

\begin{figure}
\begin{center}
\includegraphics[viewport=180 100 425 722, scale=0.42]{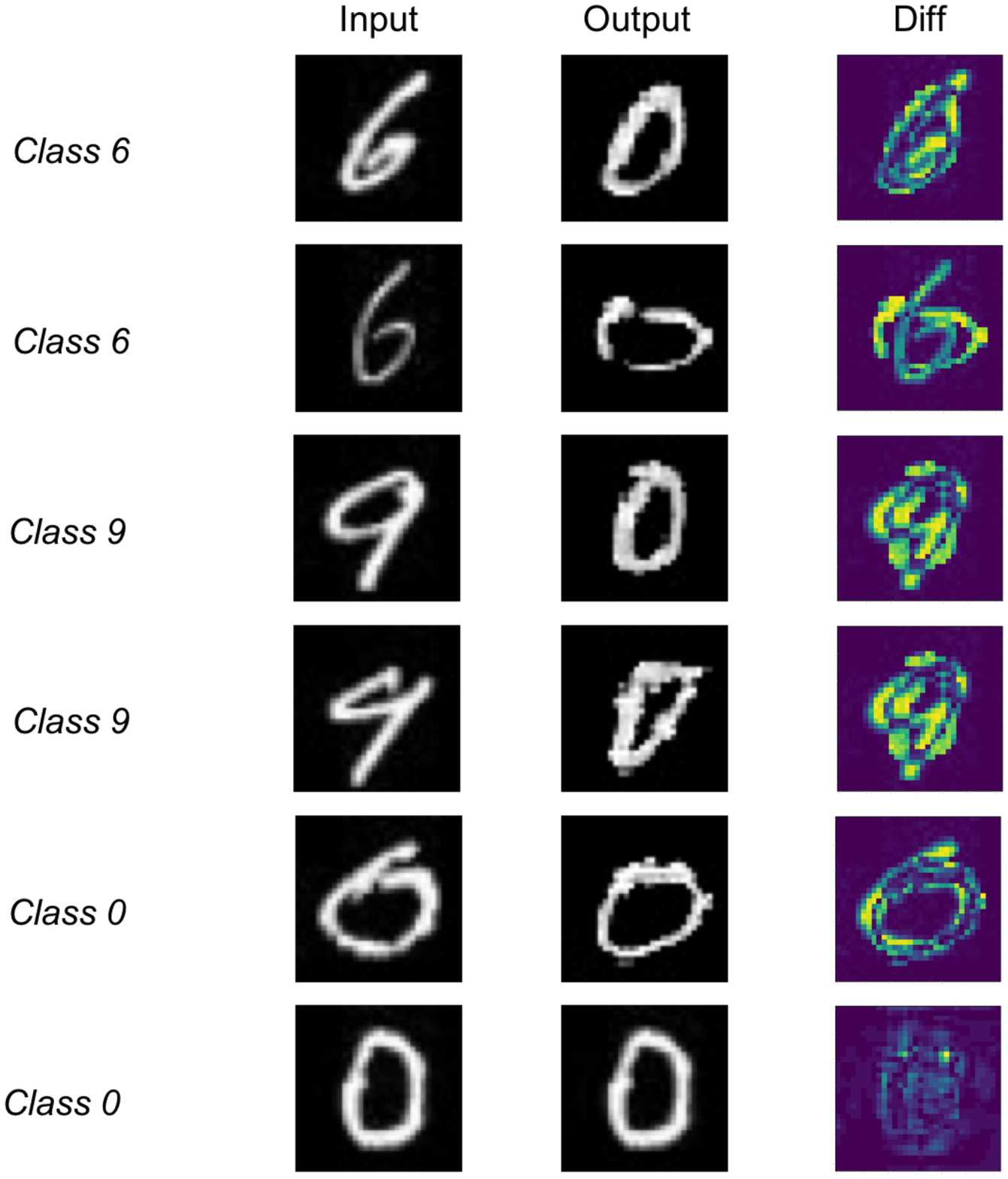}
\vspace{0.2cm} 
\caption{Experimental results on the MNIST dataset.The output images were generated by the proposed method which trained 0 as a normal class.}
\label{fig:mnist_results}
\end{center}
\end{figure}

\section{Experiments}
Our method successfully detects pixel-level micro anomalies in $1024\times{1024}$ high resolution images from the real industrial data with a high accuracy.
However, we conducted an experiment with open dataset as we cannot disclose the images due to the confidentiality of the data.
We use $\gamma = 0.1$, $\lambda = 0.1$ and $n = 500$ as each parameter of the proposed method.

\subsection{Datasets}
We use the following two datasets for experiments. 
Table\ref{table:dataset} shows the list of datasets.\\
{\bf MNIST}: This dataset\footnote{$http://yann.lecun.com/exdb/mnist/$} includes $70,000$ handwritten digits from $0$ to $9$. This dataset has a training set of $60,000$ examples and a test set of $10,000$ examples. The images included in this dataset are unified to the size of $28\times{28}$ pixels. We regard one class out of ten classes as a normal data and we consider images of other classes to be abnormal. We select a class as a normal class and allocate $70\%$ of it to the training set. In addition to that, $10\%$ from all the classes other than the normal class is added to the training data so that the training data can have both normal and abnormal class.
After training, we test with the data that was not used for learning and contains both of the classes .
This experiment is repeated for all of the $10$ classes. \\
{\bf Caltech-256}: This dataset\footnote{$http://www.vision.caltech.edu/Image_Datasets/Caltech256/$} includes $30,607$ images of $256$ object categories. Each category has at least $80$ images. 
We follow the experimental setup of Sabokrou et al\cite{sabokrou2018adversarially}.
In addition, we sample $50$ images from all outlying category images, add them to the training data and train the model with the proposed method. \\
{\bf IR-MNIST}: This dataset\footnote{ $http://ai.stanford.edu/~eadeli/publications/data/IR-MNIST.zip$} is made of MNIST dataset.
To create each sample, randomly 121 samples are selected from the MNIST dataset and are put together as a $11\times{11}$ puzzle.
Training samples are created without using any images of the digit ‘3’. Hence, ‘3’ is considered as an abnormal class.
We train the proposed method using $5000$ normal class data and $50$ abnormal class data.

\begin{table}[htbp]
  \caption{An overview of the datasets used for the experiments.}\label{table:dataset}
    \centering
    \vspace{0.2cm} 
    \begin{tabular}{lll}
    dataset name & dataset size & class\\
    \midrule
    MNIST & $70,000$ & 10 \\
    Caltech-256 & $30,607$ & 256 \\
    IR-MNIST & $6,051$ & 10 \\
    \bottomrule
    \end{tabular}
\end{table}

\begin{figure}
\begin{center}
\includegraphics[viewport=180 300 425 600, scale=0.65]{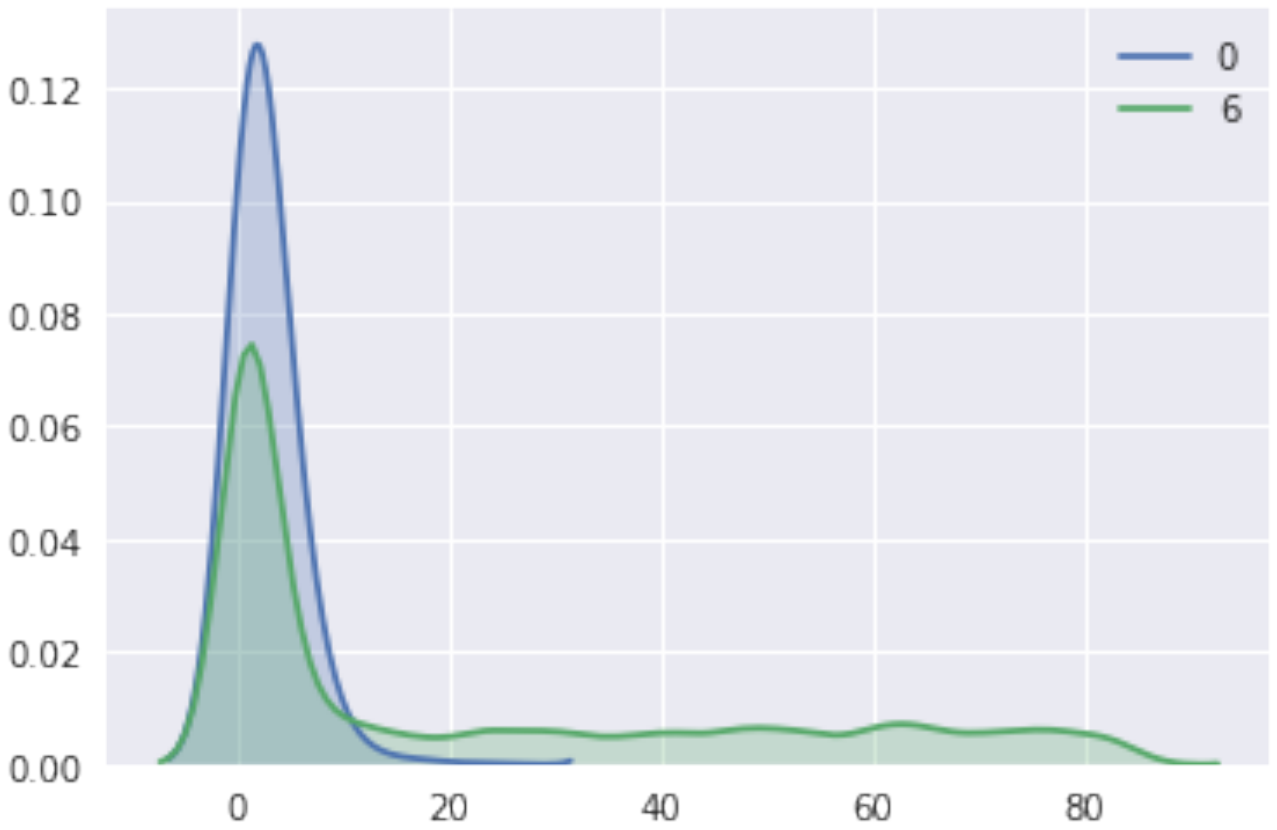}
\vspace{0.2cm} 
\caption{Distribution of pixel difference between input image and generated image.}
\label{fig:mnist_distribution}
\end{center}
\end{figure}

\begin{table*}[bthp]
\caption{Results on the Caltech-256 dataset. We used the result of Sabokrou\cite{sabokrou2018adversarially} for the comparison method other than AnoGAN\cite{schlegl2017unsupervised} and the proposed method. We used Karras et al.'s network architecture for AnoGAN and out proposed method. The metric of this table is the $F1$-score, and $N$ is the number of categories of the normal class.}\label{table:caltech_results}
\centering
\vspace{0.2cm} 
\begin{tabular}{ccccccccc}
N & CoP\cite{rahmani2017coherence} & REAPER\cite{lerman2015robust} & LRR\cite{liu2010robust} & DPCP\cite{tsakiris2015dual} & R-graph\cite{you2017provable} & Sabokrou\cite{sabokrou2018adversarially} & AnoGAN\cite{schlegl2017unsupervised} & Ours \\
\midrule
$1$ & $0.880$ & $0.808$ & $0.893$ & $0.785$ & $0.914$ & $0.928$ & $0.956$ & \mbox{\boldmath$0.977$} \\
$3$ & $0.718$ & $0.784$ & $0.671$ & $0.777$ & $0.880$ & $0.913$ & $0.915$ & \mbox{\boldmath$0.963$} \\
$5$ & $0.672$ & $0.716$ & $0.667$ & $0.715$ & $0.858$ & $0.905$ & $0.887$ & \mbox{\boldmath$0.945$} \\
\bottomrule
\end{tabular}
\end{table*}

\subsection{Network Architecture}
We use the progressive growing ~\cite{karras2018progressive} as a learning framework and further details about the network architecture and the hyper parameters can be found in Table 2 and Appendix A in the corresponding paper.

\subsection{Results}
\begin{figure*}[tbp]
\begin{center}
    \includegraphics[viewport=0 0 948 503, scale=0.5]{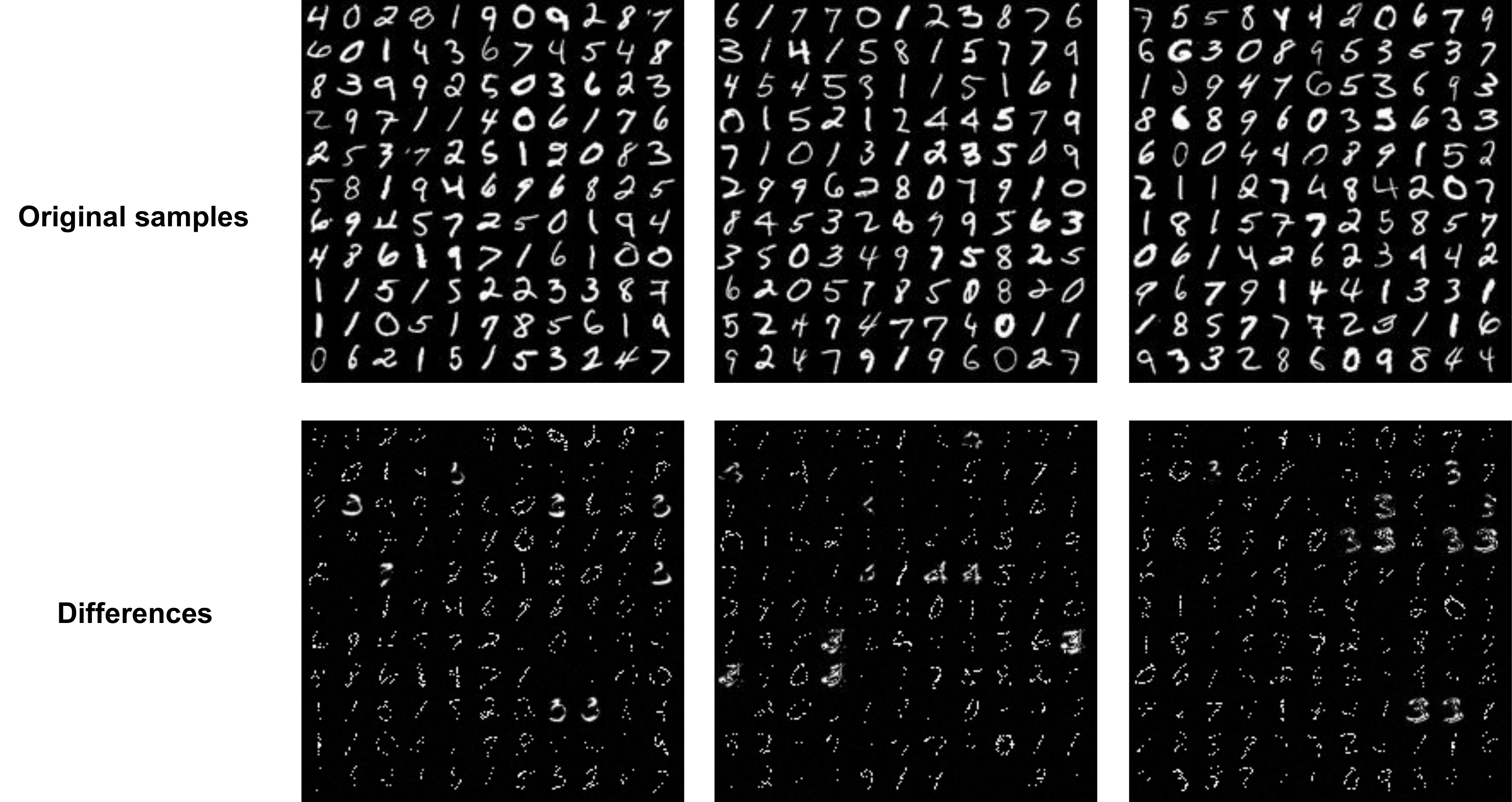}
\end{center}
\caption{Experimental results on IR-MNIST dataset. The top images are the original test images. In the test image, 3 exists unlike the training image. The bottom images are difference images between the outputs of the proposed method and the test samples.}\label{fig:ir_mnist}
\vspace{0.2cm}
\end{figure*}

We introduce some experimental results on the benchmark dataset. \\
{\bf MNIST Results}:  Figure\ref{fig:mnist_results} shows the experimental results for the MNIST dataset. 
We regarded 0 as a normal class, and 1 as the abnormal class. 
We could almost perfectly generate images belonging to the normal class at the bottom, but images of abnormal class and normal noisy class was not properly generated. 
Figure\ref{fig:mnist_distribution} shows the distribution of the pixel difference between the input image and the generated image for label 0 and label 6 using the same model.
We show the results of sampling and inferring 100 images for each class and averaging them. \\
{\bf Caltech-256 Results}: Table\ref{table:caltech_results} shows the experimental results with the Caltech-256 dataset.
We used the result of Sabokrou\cite{sabokrou2018adversarially} for the comparison method other than AnoGAN\cite{schlegl2017unsupervised} and the proposed method.
In this experiment, we evaluate all the methods based on the $F_1$-score metrics with different numbers of normal-classified categories  $n\in{\{1, 3, 5\}}$.
In all three cases, our proposed method outperformed all other methods and maintained its high performance even when the number of normal-classified categories increased, whereas AnoGAN's score dropped under such a circumstance. 
{\bf IR-MNIST Results}: Figure\ref{fig:ir_mnist} shows the experimental results with the IR-MNIST dataset. 
The top images are the original test images.
The bottom images are difference images between the outputs of the proposed method and the test samples.
In this dataset, ‘3’ does not exist in the normal classes, and it can be seen from the difference image that a large difference is obtained in this class.

\section{Conclusion}
In this paper, we proposed a method of detecting anomalies with GANs using both normal images and given abnormal images.
By distorting the data distribution and excluding the distribution of the abnormal images, the network can learn more about ideal normal data distributions. This method allows us to make more robust and accurate models to detect anomalies; our model detected smaller than 1\% of abnormal pixels in  $1024\times{1024}$ high resolution images. Due to the confidential nature of the data, we share only the results for open datasets, and further validation on various datasets is desirable.

{\small
\bibliographystyle{ieee}
\bibliography{main}
}

\end{document}